%% file: main.tex
\documentclass[10pt,twocolumn,letterpaper]{article}

\usepackage[ruled,vlined,noend,linesnumbered]{algorithm2e}

\usepackage[pagenumbers]{cvpr} 

\input{preamble}

\definecolor{cvprblue}{rgb}{0.21,0.49,0.74}
\usepackage{cuted}
\usepackage{float}
\usepackage{listings}
\usepackage{makecell}
\usepackage{tcolorbox}
\usepackage{multirow}
\usepackage{pifont}
\usepackage[pagebackref,breaklinks,colorlinks,allcolors=cvprblue]{hyperref}
\usepackage{fvextra} 

\newcommand{\cmark}{\ding{51}}%
\newcommand{\xmark}{\ding{55}}%

\renewcommand{\paragraph}[1]{\vspace{.1em}\noindent\textbf{#1}}

\title{
    SIMPACT: Simulation-Enabled Action Planning using Vision-Language Models
}

\author{
    Haowen Liu$^{*,1}$\quad Shaoxiong Yao$^{*,2}$\quad Haonan Chen$^{3}$\quad Jiawei Gao$^{3}$\\
    Jiayuan Mao$^{4,5}$\quad Jia-Bin Huang$^{1}$\quad Yilun Du$^{3}$\\
    \vspace{0.8em}
    \small{$^{1}$UMD, $^{2}$UIUC, $^{3}$Harvard, $^{4}$Amazon FAR, $^{5}$UPenn}\\
    \vspace{-44pt}
}

\begin{document}
\maketitle

\renewcommand{\thefootnote}{\fnsymbol{footnote}}
\footnotetext[1]{Equal contribution}
\input{sec/figs/teaser}

\input{sec/0_abstract} 
\input{sec/1_intro}
\input{sec/2_related_works}
\input{sec/3_method}

\input{sec/4_experiments}
\input{sec/5_conclusion}

{
    \small
    \bibliographystyle{ieeenat_fullname}
    \bibliography{main}
}

\input{sec/X_suppl}

\end{document}

%% file: preamble.tex
\newcommand{\topic}[1]{\vspace{2mm}\noindent\textbf{#1}}



%% file: sec/figs/teaser.tex
\begin{strip}
    \centering
    \includegraphics[trim={1.5cm 18.1cm 3cm 4.7cm},clip,width=\textwidth]{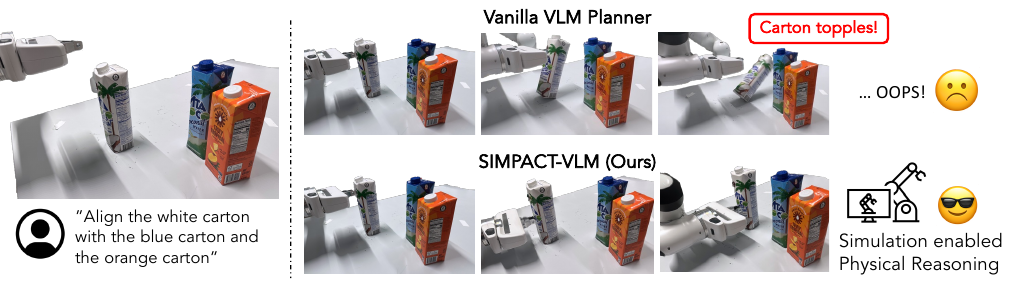}
    \vspace{-3em}
    \captionof{figure}{\textbf{Simulation-Enabled VLM Action Planning.} 
    Given a single RGB--D image and a language task description (\emph{left}), our method efficiently constructs a physics simulator that enables test-time VLM reasoning with physical grounding.
    This physically grounded reasoning allows the robot to succeed in fine-grained manipulation tasks (\emph{bottom}), outperforming a vanilla VLM planner (\emph{top}) that lacks awareness of physical dynamics. }
    \vspace{-10px}
    \label{fig:star}
\end{strip}

%% file: sec/0_abstract.tex
\begin{abstract}
Vision-Language Models (VLMs) exhibit remarkable common-sense and semantic reasoning capabilities.
However, they lack a grounded understanding of physical dynamics. 
This limitation arises from training VLMs on static internet-scale visual-language data that contain no causal interactions or action-conditioned changes.
Consequently, it remains challenging to leverage VLMs for fine-grained robotic manipulation tasks that require physical understanding, reasoning, and corresponding action planning.
To overcome this, we present \textbf{SIMPACT}, a test-time, \textbf{SIM}ulation-enabled \textbf{ACT}ion \textbf{P}lanning framework that equips VLMs with physical reasoning through simulation-in-the-loop world modeling, without requiring any additional training.
From a single RGB-D observation, SIMPACT efficiently constructs physics simulations, enabling the VLM to propose informed actions, observe simulated rollouts, and iteratively refine its reasoning.
By integrating language reasoning with physics prediction, our simulation-enabled VLM can understand contact dynamics and action outcomes in a physically grounded way. 
Our method demonstrates state-of-the-art performance on seven challenging, real-world rigid-body and deformable manipulation tasks that require fine-grained physical reasoning, outperforming existing general-purpose robotic manipulation models. 
Our results demonstrate that embedding physics understanding via efficient simulation into VLM reasoning at test time offers a promising path towards generalizable embodied intelligence. Project webpage can be found at \textcolor{blue}{\href{https://simpact-bot.github.io/}{https://simpact-bot.github.io}}.
\end{abstract}

%% file: sec/1_intro.tex
\section{Introduction}
\label{sec:intro}
General-purpose robots hold significant promise for handling complex, labor-intensive tasks in unstructured environments, but realizing this potential requires advanced scene perception and robust action planning.
Vision-Language Models (VLMs), trained on static internet-scale visual and language data, offer a promising solution by equipping robots to understand scenes and respond to diverse queries. 
These models can understand object semantics, infer task goals, and generate action descriptions aligned with human intent~\cite{openai2023gpt4,comanici2025gemini,driess2023palme}. 
However, despite their remarkable commonsense and semantic reasoning capabilities, VLMs lack a \textbf{grounded understanding of physical dynamics}. 
They can describe what to do, but often fall short in predicting how actions will unfold when executed in the physical world.

As such, VLMs have shown limited capabilities in robotic manipulation, particularly for tasks involving rich physical interactions, such as turning an object in place or carefully stacking objects.
These tasks require reasoning about how objects behave under forces and constraints, where small variations in contact or timing can lead to drastically different outcomes. 
Lacking physical understanding, VLMs often propose plans that appear reasonable in language but fail during execution.

To address this limitation, we propose a framework that augments VLMs with physical simulation rollouts as contextual feedback, enabling test-time physical reasoning for action planning. Our approach begins with a novel simulation generation pipeline that leverages pretrained visual foundation models—including segmentation, 3D generation, and pose estimation models —to efficiently build a physical simulator directly from a single-view RGB-D image.
In addition, we use VLMs to automate the setup of a multi-physics simulator, enabling it to model the behavior of both rigid and deformable objects across diverse material properties.
The resulting physical simulation characterizes intricate contact dynamics that are difficult to infer from static images and language alone, providing VLMs with physical insights for manipulation planning.

Powered by the generated simulation, we introduce a planning framework driven by VLMs' reasoning capabilities. 
Our key idea is to leverage the rich prior knowledge of VLMs to generate action sequence proposals, and to use simulated rollouts as context for the VLM to iteratively refine these proposals. 
This test-time reasoning paradigm, inspired by model-based control frameworks~\cite{williams2015model,rawlings2020model}, enables VLMs to reason not only about the world through language but also about its dynamics through simulated interaction.
By augmenting VLMs with physical simulation, our framework enables them to anticipate action consequences, evaluate predicted outcomes, and iteratively adjust their decisions at test time, without any task-specific training. 
This process unlocks significantly stronger physical reasoning, enabling more reliable and robust real-world performance than state-of-the-art general-purpose manipulation models.
In summary, this paper makes the following contributions:
\begin{itemize}[leftmargin=2em, nosep]
    \item We introduce a test-time, zero-shot framework enabling VLMs to plan physics-aware embodied actions;
    \item We present a pipeline for automatically generating multi-physics simulations from a single RGB-D observation using visual foundation models and VLM; 
    \item We propose a novel in-context learning approach for robot action generation, where physics simulation serves as context, enabling a new form of test-time reasoning in robotics.
\end{itemize}

%% file: sec/2_related_works.tex
\begin{figure*}[!ht]
    \centering
    \includegraphics[trim={3.7cm 13.2cm 1.4cm 8.35cm},clip,width=\linewidth]
    {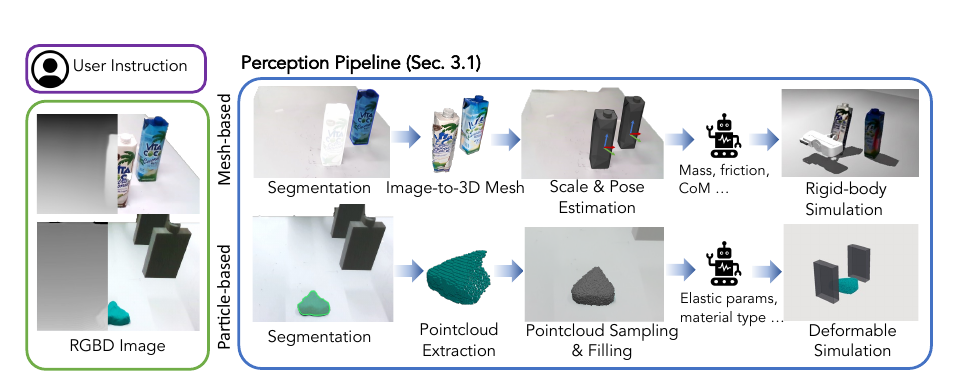}
    \vspace{-2em}
    \caption{
    \textbf{Simulation construction from a single RGBD image.} 
    Given an RGB-D image and a language task description, our pipeline automatically generates either a mesh-based simulation (\emph{top}) for rigid objects or a particle-based simulation (\emph{bottom}) for deformables. 
    After segmenting objects-of-interest via GroundedSAM2~\cite{ren2025grounded2}, we reconstruct either the 3D shape, scale, and pose of the object for rigid-body simulation, or perform dense sampling of particles within the volumes between the object surface and the table for the particle-based simulation pipeline. 
    In both cases, we prompt the VLM to infer the relevant physical parameters required for simulation.
    }
    \label{fig:simulator}
    \vspace{-15pt}
\end{figure*}

\section{Related Works}
\label{sec:related_works}
\vspace{-5pt}

\topic{Vision-Language Models for Robotics.}
VLMs excel at scene understanding and language interpretation~\cite{radford2021learning,ramesh2021zero, li2022blip,openai2023gpt4,li2023blip}, making them promising for natural language-based robot control in open-world environments~\cite{hu2024look,du2024video,chen2024spatialvlm,wang2024grounding,duan2024manipulateanything}. 
Many existing works directly fine-tune an action prediction head, i.e., vision-language-action (VLA) models, but these methods require large amounts of action-labeled training data, and their generalization ability remains limited~\cite{brohan2023rt, mu2023embodiedgpt,zawalski2024robotic,black2025pi,pi0.5}.
Other works adopt carefully designed 3D geometric representations to enable VLMs to reason about actions. 
Examples include volumetric value maps~\cite{huang2023voxposer}, motion arrows~\cite{nasiriany2024pivot}, keypoint affordances~\cite{fangandliu2024moka,yuan2024robopoint}, and keypoint constraints~\cite{huang2024copa,huang2024rekep}. 
While these spatial representations advance VLM reasoning capacity through 3D grounding, they critically lack temporal dynamics, which are essential for tasks involving physical interaction and sequential manipulation. Early works have explored using physics simulation to augment reasoning in VLMs~\cite{liu2023minds}, and physical grounding for VLMs has also been investigated~\cite{gao2024physically}. 
However, these efforts focus on question answering rather than the more challenging task of manipulation planning, which requires a robot to generate and refine continuous action sequences.

\looseness=-1
\topic{Model-based Planning in Robotics.}
Model-based planning has long been studied in robotics as a generalizable way to automatically synthesize long-horizon, complex action sequences~\cite{lavalle2006planning,mordatch2012cio,garrett2021tampsurvey,howell2022mjmpc, kang2024fast, kang2025global}.
However, existing planning frameworks face limitations in open-world settings, where we need to build dynamic models from real-world perception and plan long-horizon actions from language instructions.
With advances in deep learning, neural dynamics models have been developed to capture physical dynamics through image-space prediction~\cite{finn2017deepvisualforesight, yang2024learning, du2024video}, latent-space dynamics modeling~\cite{agrawal2016learning, neurips2018worldmodel}, and structured world representations~\cite{chen2023predicting, zhang2024adaptigraph, zhang2024particle, chen2025bimanual}. 
To improve planning efficiency, methods have been proposed to learn sampling distributions~\cite{qureshi2019motion,power2024tro} or to increase optimization efficiency using energy-based approaches~\cite{du2019modelbp, janner2022planningwd}.
Nevertheless, these extensions still require training within a specific problem domain.

We argue that existing works do not fully address the open-world manipulation challenge.
Task-specific models struggle with the diversity of real-world scenes.
In contrast, pre-trained VLMs offer general scene understanding and reasoning capabilities, so we leverage them to support each component of our framework.

Our work also advances the construction of simulations from real-world observations.
Compared to real-to-sim-to-real approaches such as digital twins~\cite{patel2025real,wang2025phys2real,jiang2025phystwin,xia2018gibson,torne2024reconciling,han2026re3sim} and cousin creation~\cite{dai2024automated}, our approach constructs simulations more efficiently from a single-view RGB-D image.
The recent method Prompting-with-the-Future~\cite{ning2025prompting} uses rigid-body simulation and a VLM solely as a reward signal in a model-predictive-control setup.
In contrast, our method integrates multi-physics simulation and exploits VLMs for both informed action sampling and in-context learning optimization, resulting in substantial performance gains as demonstrated by our experiments.

%% file: sec/3_method.tex
\section{Method}
\label{sec:method}

Our framework enables zero-shot robotic manipulation action generation from a single RGB-D image input $I_0$ and natural language instruction $\ell_{\text{task}}$ and outputs robot action sequence $\mathbf{a} = \{a_t\}_{1 \leq t \leq T}$, where $a_t \in \mathrm{SE}(3) \times \mathbb{R}$, defining end-effector pose and gripper open width. 
For each task, the natural-language specification $\ell_{\text{task}}$ defines the task requirements, along with potential success and failure conditions, to guide the VLM in proposing plausible actions. 

\begin{figure}
    \centering
    \includegraphics[trim={4.8cm 14.6cm 6.8cm 7.8cm},clip,width=\linewidth]
    {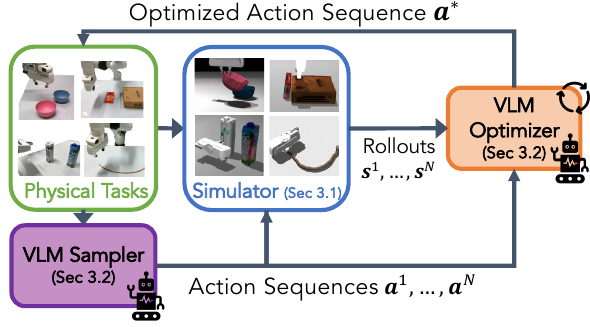}
    \vspace{-2em}
    \caption{\textbf{Method overview}. Our method first instantiates a physics simulator given the real-world scene. Next, a VLM-based action sampler and optimizer iteratively refine the action sequence towards task success using simulated rollouts as context. The final optimized actions are then executed in the real world.}
    \label{fig:overview}
\end{figure}

Our simulation-enabled VLM planning framework operates as illustrated in Fig.~\ref{fig:overview}.
First, we construct a physical simulator $\textsc{Sim}$ using an automated perception pipeline that reconstructs complete 3D geometries and configures appropriate simulation parameters as shown in Fig.~\ref{fig:simulator}.
Next, we instantiate a manipulation planner that integrates the simulator with a VLM as its core reasoning module. 
The planner begins by generating a scene context from an initial visual observation, which is augmented with robot proprioceptive data and object states. 
Based on this context and prior knowledge, the VLM proposes action sequences, which are evaluated through simulation rollouts. 
The resulting visual observations and object states from each rollout are then fed back to the VLM as additional context for iterative refinement. 
This process continues until a rollout is validated as successful. Finally, the optimized action sequence is executed as end-effector commands on the real robot system. 

\subsection{Simulation Construction}

Our approach employs a physics-based simulator to predict the consequences of actions for manipulation planning. 
The simulation follows the discrete-time state transition:
\begin{equation}
    s_t = \textsc{Sim}(s_{t-1}, a_t; \theta)
\end{equation}
where $s_t$ denotes the state at time step $t$, $a_t$ represents the applied action, and $\theta$ comprises time-invariant simulation parameters. 
The state space captures all task-relevant information: rigid objects are represented by a 6DoF pose in $\mathrm{SE}(3)$, while deformable objects are described by $N$ particle positions in $\mathbb{R}^{3 \times N}$.
We initialize the state as $s_0$, assuming objects remain static prior to interaction, and construct parameters via $\theta = \mathrm{CreateSim}(I_0)$ from the initial RGBD image $I_0$.
Here, the simulation parameters are defined as $\theta = (\theta_{\text{geom}}, \theta_{\text{phys}})$, where $\theta_{\text{geom}}$ specifies the object shape and pose, and $\theta_{\text{phys}}$ characterizes its mechanical properties.

Our geometry pipeline begins by prompting a VLM to generate object labels based on the user's instructions, as shown in Fig.~\ref{fig:simulator}. 
We first apply a pretrained segmentation model, GroundedSAM2~\cite{ren2024grounded,ren2025grounded2}, to segment each identified object in $I_0$.
We prompt the VLM to automatically select different physics engines based on object characteristics: MuJoCo~\cite{mujoco} for rigid bodies, a variant of the projective dynamics~\cite{bouaziz2014pd} solver for stiff deformable objects that ensures numerical stability, and the Material Point Method~\cite{jiang2016mpm} solver for soft objects to handle potential topological changes. 
We automate physical parameters $\theta_{\text{phys}}$ inference by prompting the VLM to leverage its commonsense reasoning for plausible predictions, following prior works~\cite{xie2024deligrasp,xia2024video2game,chen2025physgen3d}, with more details in Suppl. Sec.~\ref{subsec:sim_construct}.

\looseness=-1
\paragraph{Mesh-based Rigid Body Simulation.}
For rigid bodies, we define the geometry parameters as $\theta_{\text{geom}} = \{(\mathcal{M}_i, X_i)\}_{i=1}^{N_{\text{obj}}}$, 
where $\mathcal{M}_i$ denotes the triangle mesh and $X_i$ represents the initial 6DoF pose of object $i$. 
Using the segmented RGB image, we reconstruct complete triangle meshes for each object using a pretrained image-to-3D model~\cite{hunyuan3d2025hunyuan3d}, denoted as the unscaled mesh $\hat{\mathcal{M}}_i$. 
Each reconstructed mesh is then centered and scaled according to the size of its corresponding real-world bounding box obtained from point cloud segmentation, yielding $\mathcal{M}_i = \alpha_i (\hat{\mathcal{M}}_i - \mathbf{\beta}_i)$, where $\alpha_i$ denotes the ratio between the diagonal length of the real-world bounding box and that of the unscaled mesh, and $\mathbf{\beta}_i$ represents the 3D centroid of the unscaled mesh
Finally, we estimate the 6DoF pose $X_i$ for each object using its triangle mesh $\mathcal{M}_i$ in model-based mode and the RGB-D observation $I_0$, employing FoundationPose~\cite{wen2024foundationpose}.
The physical parameters $\theta_{\text{phys}}$ include mass, friction, and the center of mass for each rigid body.

\paragraph{Particle-based Deformable Object Simulation.}
For deformable objects, we define $\theta_{\text{geom}} = \{P_i\}_{i=1}^{N_{\text{obj}}}$, where each $P_i \subset \mathbb{R}^3$ denotes the point set representing object $i$.
We first back-project the segmented object mask from the depth image to obtain 3D surface points. 
To construct the full particle representation, we uniformly sample points within the volume bounded by 
the object surface and the supporting table surface, as illustrated in the bottom row of Fig.~\ref{fig:simulator}.
Deformable bodies have $\theta_{\text{phys}}$ defined by elasticity and plasticity parameters; see Suppl. Sec.~\ref{subsec:sim_construct} for details.

\subsection{Action Planning via Simulation-enabled VLM}

\begin{algorithm}[t]
\caption{Action Planning Algorithm}
\label{alg:planning}
\footnotesize
\SetAlgoLined
\textbf{Input:} $\textrm{VLM}$, $\textsc{Sim}$, $I_0$, $\ell_{\text{task}}$, $s_0$\;
$\mathcal{A} = \emptyset,\; \mathcal{S} = \emptyset$ \tcp*{Action sequences and rollouts}
\tcp{Initial action sampling and simulation}
\For{$k = 1..K$}{ 
    $\mathcal{A} \gets \mathcal{A} \cup \{\mathbf{a}^i \gets \textsc{Sample}\big(I_0, \ell_{\text{task}}, s_0;\textsc{VLM}\big)\}$\;
    $\mathcal{S} \gets \mathcal{S} \cup \{\mathbf{s}^i \gets \textsc{SimRollout}\big(s_0, \mathbf{a}^i;\textsc{Sim}\big)\}$\;
}

\tcp{Iterative action optimization}
\For{$k = K\!+\!1$ \KwTo $K_{\max}$}{
    $\mathbf{a}^{k} \gets 
        \textsc{Optimize} \big(
            \mathcal{A}, \mathcal{S}, \ell_{\text{task}}; \textrm{VLM}
        \big)$\;

    $\mathbf{s}^{k} \gets \textsc{SimRollout}\big(s_0, \mathbf{a}^k;\textsc{Sim}\big)$\;

    \If{$\textsc{TaskSuccess}\big(\mathbf{s}^{k};\textsc{VLM}\big)$}{
        \textbf{break} \tcp*{Stop when successful}
    }
    \Else{
        $\mathcal{A} \gets \mathcal{A} \cup \{\mathbf{a}^{k}\}, \; \mathcal{S} \gets \mathcal{S} \cup \{\mathbf{s}^{k}\}$\;
    }
}
\Return{$\mathbf{a}^{k}$} 
\end{algorithm}

Given the constructed simulator $\textsc{Sim}$, our action planning framework follows an iterative refinement process, as outlined in Fig.~\ref{fig:overview}.
As shown in Alg.~\ref{alg:planning}, our planner takes as input the initial RGB-D observation $I_0$, the initial simulator state $s_0$, task description $\ell_{\text{task}}$, VLM, and $\textsc{Sim}$. 
The planner begins by sampling an initial set of action sequences $\mathcal{A}$ from the VLM prior. For each action sequence $\mathbf{a}^i \in \mathcal{A}$, the \textsc{SimRollout} procedure iteratively applies each action $a_t^i$ and uses the \textsc{Sim} function to obtain the next state $s_{t+1}^i$, adding simulation rollouts $\mathbf{s}^i \in \mathcal{S}$.

\begin{figure*}[t]
    \centering
    \includegraphics[trim={2.9cm 13.3cm 3cm 11.6cm},clip,width=\linewidth]
    {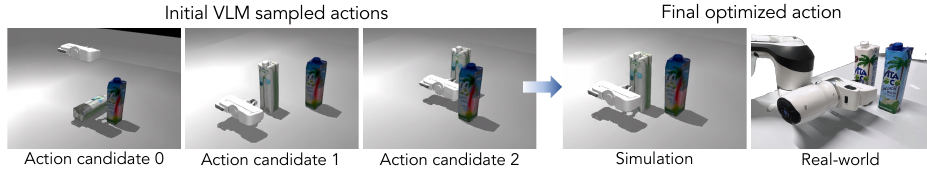}
    \vspace{-2.2em}
    \caption{\textbf{Action optimization process.} We show a representative example from the \emph{non-toppling push} task. The left three images show simulation rollouts from initial VLM-sampled action sequence proposals, all of which fail due to insufficient/overshooting push, or because the bottle topples. From these proposals, the VLM optimizer reasons a non-trivial action update that pushes the bottle for the correct distance without toppling in both simulation and real-world execution.}
    \label{fig:action_optimize}
    
\end{figure*}

\setlength{\tabcolsep}{5mm}{
\begin{table*}[t]
    \setlength{\tabcolsep}{3pt}
    \footnotesize
    \centering
    \caption{\textbf{Definition of tasks}. For each manipulation task, we list the corresponding instruction and success criteria.}
    \vspace{-5pt}
    \begin{tabular}{lp{7.8cm} p{6.8cm}}  
    \toprule
    Tasks & Instruction & Success Condition \\
    \midrule
    Non-toppling push 
        & Push the white carton forward to align horizontally with the others.
        & The bottle does not topple and aligns within $\pm$2 cm. \\

    Bowl stacking
        & Grasp the pink bowl at its edge and stack it with the blue bowl.
        & The pink bowl stably lies inside the blue bowl. \\

    Pivoting
        & Make the red pocky box lean vertically against the brown box.
        & The red pocky box reaches vertical pose. \\

    Shape rope
        & Grab the free end of the rope and arrange the rope to a \textbf{U} shape.
        & The deformed rope has an opening ratio in $[0.5, 2.0]$. \\

    Shape playdoh
        & Squeeze the Play-Doh to a square shape with equal sides.
        & Two sides of the Play-Doh have a ratio within 1.5. \\

    Avoid obstacle 
        & Push the orange bottle around the box to the image’s bottom-right.
        & The orange bottle reaches the other side without colliding. \\
    Sweeping 
        & Sweep the coffee beans into the purple box. & All beans are pushed into the box. \\
    \bottomrule
    \end{tabular}
    \label{tab:tasks_summary}
    \vspace{-2.2em}
\end{table*}
}

After initialization, each iteration proceeds as follows. Using both $\mathcal{A}$ and $\mathcal{S}$, a VLM-based optimizer refines the proposed action sequences and produces a new action sequence $\mathbf{a}^k$. Based on the simulated rollout $\mathbf{s}^k$, the VLM model then evaluates whether $\mathbf{a}^k$ achieves the task goal. If successful, the corresponding action sequence $\mathbf{a}^k$ is executed on the real robot, and the process terminates. Otherwise, the planner proceeds to another round of action optimization by adding a newly generated  $\mathbf{a}^{k}$ to $\mathcal{A}$ and $\mathbf{s}^{k}$ to $\mathcal{S}$, until either a successful plan is found or the maximum iteration limit $K_{\max}$ is reached.

At the heart of our planning framework is the VLM, which uses its pretrained knowledge to instantiate the $\textsc{Sample}$, $\textsc{Optimize}$, and $\textsc{TaskSuccess}$ modules.
For each role, we define a corresponding system prompt $\ell_{*}$, where $*$ denotes $\text{sample}$, $\text{opt}$, or $\text{eval}$, specifying the function that the VLM performs.

\paragraph{VLMs for Action Proposal Generation.}
To instantiate $\textsc{Sample}$ using a VLM, we build upon two key ideas:  
(1) constructing an informed contextual description of the environment, and  
(2) leveraging hierarchical action generation.  

\noindent \textit{(1) Contextual representation.}  
We begin by constructing a comprehensive context that includes the initial visual observation $I_0$ and the robot’s proprioceptive state. 
For the manipulated objects, we further incorporate their 6-DoF poses along with key geometric attributes, such as bounding box dimensions. Full details are shown in Suppl. Sec.~\ref{subsec:action_details}.

\noindent \textit{(2) Hierarchical action generation.}
Directly prompting a VLM to generate continuous 6-DoF end-effector poses for long-horizon tasks is challenging, as such representations lack clear semantic meaning and are difficult for VLMs to reason about. 
In contrast, we find that VLMs are highly effective at producing high-level symbolic action sequences that align with patterns seen in their pretraining data. 
Accordingly, we define a compact set of symbolic actions e.g., \texttt{MOVE}, \texttt{GRASP}, and \texttt{RELEASE}, to better exploit the semantic reasoning capabilities of VLMs. 
Each symbolic action is further parameterized by continuous control variables, enabling fine-grained and precise motion execution.

Formally, we represent a high-level action at time $t$ as $\textsc{a}_t = (\tau_t, u_t)$
where $\tau_t$ denotes the high-level action type and $u_t$ represents the continuous control parameters.
A deterministic mapping, $\textsc{Action2Pose}$, translates a sequence of high-level actions into continuous 6-DoF control trajectories. 
Within the $\textsc{Sample}$ function, let $\textsc{a}^i$ denote the $i$-th high-level action sequence. 
The VLM then samples an action sequence as
\begin{equation*}
    \mathbf{a}^i = \textsc{Action2Pose}\big(\textsc{a}^i = \textrm{VLM}(I_0, \ell_{\text{task}}, s_0; \ell_{\text{sample}})\big).
    \label{eq:sample_action}
\end{equation*}
where $\ell_{\text{sample}}$ denotes the system prompt that specifies the sampling behavior of the VLM (refer to Suppl. Sec.~\ref{subsec:action_details} for details).

\paragraph{VLMs for Action Optimization.}
Given sampled action sequences $\mathcal{A}=\{\mathbf{a}^i\}_{i=1}^K$, we first perform simulation rollouts to obtain their corresponding state trajectories $\mathcal{S}=\{\mathbf{s}^i\}_{i=1}^K$.
Next, we instantiate the $\textsc{Optimize}$ function using the VLM via in-context learning. 
For each action sequence, we construct an optimization context $c^i$ by subsampling time steps and gathering intermediate information. 
In particular, at each selected time step $t$, we render a simulator observation image $I_t^i$ and include the numerical action $a_t^i$ and state $s_t^i$ in the context. This provides the VLM with both visual and state-based evidence to guide optimization.
\begin{equation}
    \mathbf{a}^{k} = \mathrm{VLM}(c^1,...,c^{K};\, \ell_{\text{opt}})
\end{equation}
This optimization process is not restricted to local updates as in numerical optimizers; instead, it can perform reasoning and even learn from all failure examples. 
Fig.~\ref{fig:action_optimize} illustrates a case where the VLM-based optimizer learns from failed attempts, produces a successful action sequence through its internal reasoning.

\paragraph{VLMs for Success Evaluation.}
Given the simulation rollout $s^{k}$, we render the final simulation state and extract both an observation image $I_T^{k}$ and the simulator state $s_T^{k}$. These are used as contextual inputs for the VLM to assess whether the task is successfully completed. The evaluation is formulated as
\begin{equation*}
    \textsc{TaskSuccess}(\mathbf{s}^{k}) = \mathrm{VLM}(I_T^{k}, s_T^{k}, \ell_{\text{task}}; \ell_{\text{eval}}).
\end{equation*}
If the VLM determines that the proposed action sequence achieves the task objective, the sequence is executed in the real environment. 
Otherwise, the system continues to optimize actions until the iteration limit is reached.

%% file: sec/4_experiments.tex
\begin{table*}[t]
    \centering
    \caption{\textbf{Success rates of our method and baselines.} For each task, we run 10 trials per method. Our approach consistently achieves a substantially higher success rate than baselines, highlighting the effectiveness of simulation-enabled VLMs for action planning.
    }
    \vspace{-5pt}
    \small
    \setlength{\tabcolsep}{4pt}
    \begin{tabular}{lcccccccc}
        \toprule
        Method & Non-toppling push & Bowl stacking & Pivoting & Shape rope & Shape dough & Avoid obstacle & Sweeping \\
        \midrule
        $\pi_{0.5}$\cite{pi0.5} & 0\% & 0\% & 0\% & 0\% & 0\% & 0\% & 0\% \\
        VoxPoser\cite{huang2023voxposer} & 0\% & 20\% & 0\% & 0\% & 0\% & 0\% & 20\%\\
        MOKA\cite{fangandliu2024moka} & 0\% & 10\% & 0\% & 20\% & 0\% & 0\% & 0\% \\
        Ours & \textbf{80\%} & \textbf{60\%} & \textbf{40\%} & \textbf{90\%} & \textbf{80\%} & \textbf{80\%} & \textbf{70\%} \\
        \bottomrule
    \end{tabular}
    \vspace{-1em}
    \label{tab:main_results}
\end{table*}

\begin{figure*}[tbp]
    \centering
    \includegraphics[trim={1.8cm 6.6cm 1.2cm 9.3cm},clip,width=\linewidth]
    {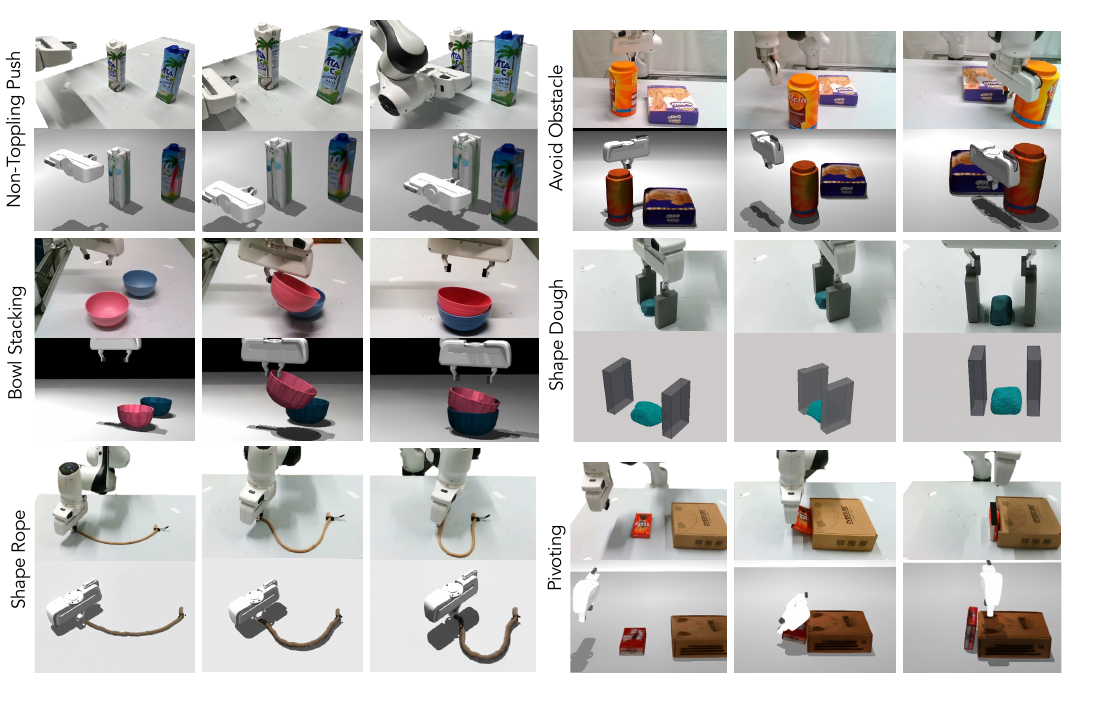}
    \vspace{-3.4em}
    \caption{\textbf{Qualitative results.} 
    The figure shows the initial state, execution progress, and final state for six tasks in both the real world (top) and the simulation (bottom). 
    By leveraging VLM's powerful generalization, rendered simulation images can guide VLM's test-time reasoning for action planning despite the visual sim2real gap. Fig.~\ref{fig:qualitative_remaining} shows the results of the remaining task.}
    \label{fig:qualitative}
    \vspace{-10pt}
\end{figure*}

\section{Experiments}

To evaluate the effectiveness of our framework, we design seven challenging, real-world, physics-aware, fine-grained manipulation tasks. 
We assess whether our method enables zero-shot planning on these tasks, comparing it against other state-of-the-art zero-shot methods. 
We validate our design choices through systematic ablation studies. 

\subsection{Experimental Setup}
We evaluate our system using a Franka Research 3 robot arm with a parallel-jaw gripper. For the Play-Doh manipulation task, we use a custom 3D-printed end effector to achieve a sufficiently large contact area. 
A single calibrated Intel RealSense D435i RGBD camera is used.

\paragraph{Tasks and Metrics.}
We design diverse tasks requiring fine-grained, physics-aware manipulation planning. 
The objects span rigid bodies (cartons, bowls, boxes) to deformable materials (rope, Play-Doh), enabling evaluation across different physical properties and manipulation strategies, including pushing, grasping, pivoting, squeezing, and sweeping.
Success rate is our primary evaluation metric. 
Task instructions and success criteria are detailed in Table~\ref{tab:tasks_summary}. 

\begin{table*}[t]
\centering
\caption{\textbf{Ablation.} Success rates (\%) over 10 trials for each task after removing each component of our method. 
Results demonstrate the importance of VLM-conditioned sampling and the VLM's simulation-enabled test-time reasoning capabilities.
}
\label{tab:ablation}
\vspace{-5pt}
\small
\setlength{\tabcolsep}{3.5pt}
\begin{tabular}{lcccccccc}
\toprule
Method & Non-toppling  push & Bowl stacking & Pivoting & Shape rope & Shape dough & Avoid obstacle & Sweeping \\
\midrule
w/o VLM sampler  & 0\% & 10\% & 0\% & 0\% & 0\% & 0\% & 20\% \\ 
w/o simulation rollout & 20\% & 0\% & 0\% & 30\% & 30\% & 0\% & 20\% \\
w/o VLM optimizer & 30\% & 50\% & 30\% & 40\% & 80\% & 20\% & 70\% \\
Ours & \textbf{80\%} & \textbf{60\%} & \textbf{40\%} & \textbf{90\%} & \textbf{80\%} & \textbf{80\%} & \textbf{70\%} \\
\bottomrule
\end{tabular}
\vspace{-0.5em}
\end{table*}

\begin{figure*}[t]
    \centering
    \includegraphics[trim={1.6cm 9.9cm 2.8cm 11.1cm},clip,width=\linewidth]
    {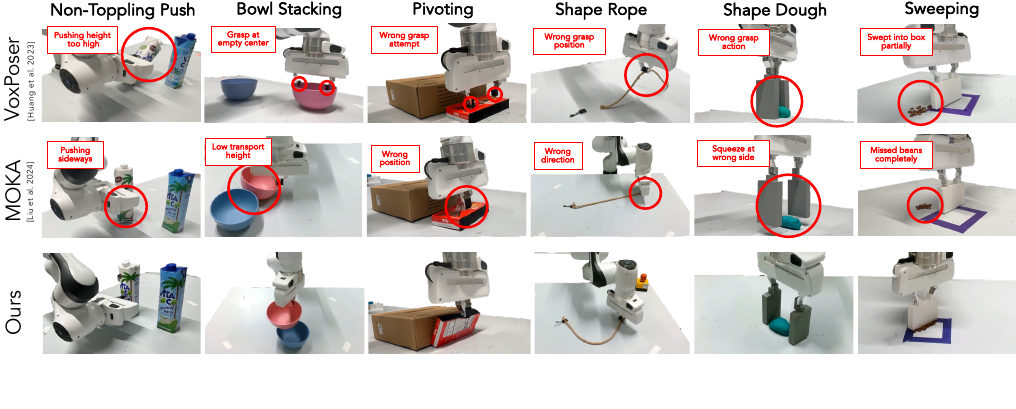}
    \vspace{-4.4em}
    \caption{\textbf{Qualitative comparison with baseline methods.}
    We show representative failures from baseline methods that lack simulation-enabled reasoning. These methods often choose incorrect action parameters, using improper pushing heights that cause toppling, or attempting to grasp the bowl at its center. Pivoting tasks fail because the baselines do not maintain contact with the box’s side face. For the rope task, baselines place the rope in the wrong direction due to missing deformation reasoning; for the dough-shaping task, baselines fail to plan the perpendicular squeezes needed to form a square; for the sweeping task, baselines fail to push all beans into the box.}
    \label{fig:comparison}
    \vspace{-1em}
\end{figure*}

\paragraph{Baselines.}
We compare our approach against the following baselines: 
(1) \emph{VLA models} that are trained on large-scale robot action datasets to directly predict joint velocities from visual observations and language instructions. 
We use $\pi_{0.5}$~\cite{pi0.5}, a recent open-source VLA model pretrained on a large robot manipulation dataset, as a representative baseline.
(2) \emph{VLM-based methods} that leverage geometric representations to augment VLM for manipulation planning. We compare against VoxPoser~\cite{huang2023voxposer}, which uses volumetric value maps to represent spatial affordances in 3D, and MOKA~\cite{fangandliu2024moka}, which predicts keypoints and affordance regions to generate manipulation actions. 
For our pushing and squeezing scenarios, we extend MOKA, which originally supports only grasping, to represent contact location with a target contact point and infer contact direction from pre- and post-contact positions.

\paragraph{Implementation details.}
For simulation, we implement the projective dynamics variant solver using PyTorch~\cite{paszke2019pytorch} and the MPM simulator using Warp~\cite{warp2022}. 
We use Google Gemini~2.5 Pro as the default VLM~\cite{comanici2025gemini} and generate $K = 10$ initial action proposals, setting $K_{\max} = 13$, corresponding to a maximum of 3 action-optimization iterations.

\subsection{Results}
Table~\ref{tab:main_results} shows the success rates of our method compared to baseline approaches. 
Overall, our method consistently outperforms baseline methods across all evaluated tasks, highlighting its strong performance on challenging, physics-aware, fine-grained manipulation tasks. Fig.~\ref{fig:qualitative} shows simulation and real-world rollouts of six of our seven tasks. 

From the table, the VLA model $\pi_{0.5}$ consistently fails on all tasks. 
While we observe that $\pi_{0.5}$ can sometimes generate actions that approach the target object, it fails to complete the manipulation. 
This is because while VLA models can perform zero-shot on tasks similar to those seen during training, they generalize poorly to out-of-domain, challenging tasks used in our experiment. 
VLM-based methods, VoxPoser and MOKA, leveraging VLM's strong scene-understanding and reasoning capabilities, achieve non-zero success rates on tasks such as \emph{bowl stacking}, \emph{shape rope} and \emph{sweeping}. 
However, they struggle with tasks that require precise action planning, where small errors, such as pushing the wrong part of an object (in \emph{non-toppling push}) or squeezing an incorrect region of deformable materials (in \emph{shape dough}) lead to failures, as shown in Fig.~\ref{fig:comparison}. 
In contrast, our method integrates simulation-enabled reasoning with VLM, enabling the robot to iteratively refine its action plan using simulation rollouts as context. 
This enables the system to identify and avoid physically unstable or ineffective strategies. 
For example, in \emph{non-toppling push}, the simulation shows that pushing near the top of the carton would cause toppling, so the system adapts by pushing from a more stable point, as shown in Fig.~\ref{fig:action_optimize}. 

\subsection{Ablation study}

We consider three ablated variants of our method.  
(1) \emph{Removing the VLM sampler}: To assess the importance of VLM-guided action sampling, we replace them with uninformed sampling from a Gaussian distribution over gripper pose deltas. 
To ensure fairness, we increase the sample size by 5$\times$. 
This variant resembles the Prompting-with-the-Future approach~\cite{ning2025prompting}, but uses a VLM-based optimizer rather than the cross-entropy method (CEM). 
The VLM optimizer is more effective due to its reasoning capability, as shown in Fig.~\ref{fig:action_optimize}, rather than being limited to the local action updates assumed by CEM.
(2) \emph{Removing simulation rollout context}: 
We evaluate whether current VLMs can reason effectively without simulation rollouts. 
Following a proposer-verifier structure, the VLM generates and evaluates multiple action proposals using only its internal reasoning.  
(3) \emph{Removing the VLM optimizer}: We disable iterative refinement and let the VLM select the best action from the initial proposals based on simulation outcomes, testing whether a naive optimization process is sufficient.

From Table~\ref{tab:ablation}, removing the VLM sampling module causes a significant performance drop. 
For fine-grained manipulation tasks, purely random sampling often yields actions far from feasible solutions, providing no useful guidance for subsequent VLM reasoning. 
This underscores the importance of VLM-conditioned action sampling in generating reasonable action proposals. 
Removing simulation-rollout validation also substantially degrades performance, particularly in tasks such as \emph{bowl stacking}, \emph{pivoting}, and \emph{avoiding obstacles}. 
This indicates that language-based reasoning without physical grounding cannot reliably infer successful action. 
However, the variant still outperforms baseline methods, largely due to the hierarchical action sampling strategy introduced in Sec.~\ref{sec:method}. 
Finally, disabling the VLM optimizer also results in a performance decrease. 
This decline is especially pronounced in tasks such as \emph{non-toppling pushing}, \emph{shape rope}, and \emph{avoid obstacle}, where the initial VLM-generated samples are often inadequate for completing the tasks and require iterative refinement. 

\begin{table}[tbp]
\centering
\setlength{\tabcolsep}{4pt}
\caption{\textbf{Sampling length ablation.} Success rates (\%) over 10 trials varying numbers of in-context examples for tasks \emph{non-toppling push}, \emph{bowl stacking}, \emph{shape rope}.
}
\label{tab:length_main}
\vspace{-5pt}
\small
\begin{tabular}{lccccc}
\toprule
\#Samples & Non-toppling push & Bowl stacking & Shape rope \\
\midrule
3 samples  & 50\% & 50\% & 40\% \\
10 samples & 80\% & 60\% & 90\% \\
20 samples & 90\% & 20\% & 80\% \\
\bottomrule
\end{tabular}
\vspace{-0.3em}
\end{table}

In addition, we investigate how the number of initial VLM-sampled actions ($K$ in Alg.~\ref{alg:planning}) affects performance, reporting success rates over 10 trials in Table~\ref{tab:length_main}: using only 3 samples degrades performance, as limited rollouts fail to provide sufficient task information, leading to poor optimization; Increasing samples also does not always help, as performance drops for \emph{bowl stacking} and \emph{shape rope} when increasing samples from 10 to 20, particularly for task \emph{bowl stacking}. This likely occurs because longer contexts reduce VLM reasoning efficacy~\cite{wang2025mmlongbench}, which could be mitigated by pre-selecting informative samples. In contrast, \emph{non-toppling push} slightly improves, as its shorter action horizon limits context growth.

\subsection{Failure Case Analysis}
\label{sec:failure_analysis}
\begin{figure}[t]
    \centering
    \includegraphics[trim={1.6cm 3cm 2.8cm 3.3cm},clip,width=0.7\linewidth]{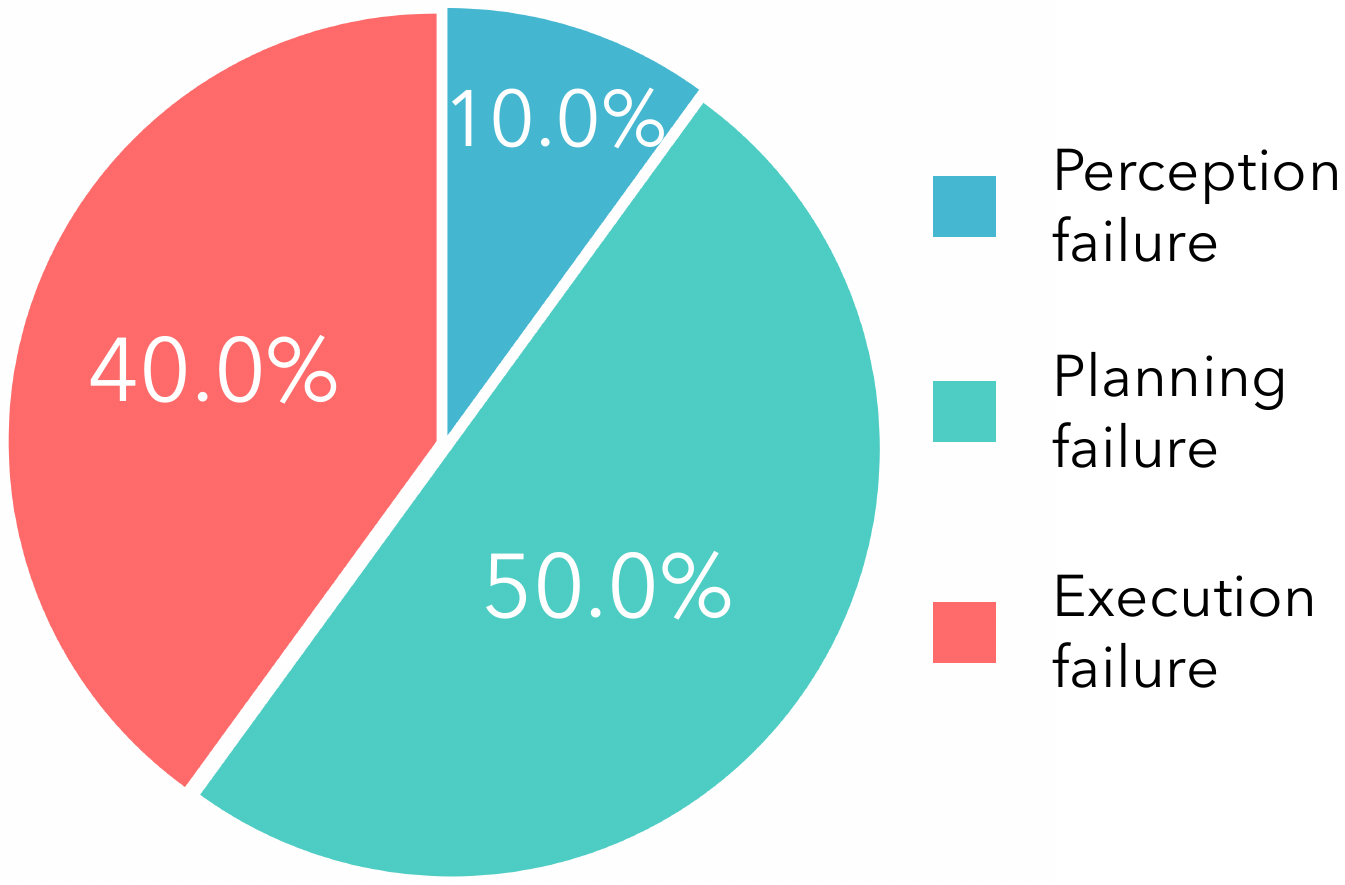}
    \vspace{-6pt}
    \caption{\textbf{Failure case decomposition graph.} 
    Failures are categorized as perception, planning, or execution.
    }
    \label{fig:failure_analysis}
\end{figure}

Fig.~\ref{fig:failure_analysis} shows the failure distribution across tasks, categorized into three failure types: perception, planning, and execution failures. 
\emph{Perception failures} mainly stem from errors in single-view 3D reconstruction, which could be reduced by using better image-to-3D models or changing the observation view.
\emph{Planning failures} occur when the robot fails to generate a feasible action sequence even after multiple rounds of action optimization. These are the most frequent failure cases, especially in the pivoting task, where finding a successful action sequence is particularly challenging.
\emph{Execution failures} arise when kinematic or dynamic discrepancies between simulation and reality cause actions that succeed in simulation to fail in real-world execution.

\subsection{Limitations}

There are several limitations to our method.
First, simulation quality depends heavily on the underlying image-to-3D reconstruction in the rigid-body case, where single-view methods struggle with occlusions. Incorporating inpainting or generative 3D completion models~\cite{lugmayr2022repaint} may help alleviate this issue. Recent works on articulated object reconstruction~\cite{chen2024urdformer,liu2025singapo} can also be integrated into our framework.
Next, since we estimate physical parameters via VLM prompting, inaccuracies can lead to sim-to-real discrepancies and affect downstream planning. Integrating system identification using real-world interaction data could help refine these estimates.
Finally, our current system performs only open-loop execution, making it vulnerable to compounding errors. We include an optional replanning mechanism for recovery that updates the simulator using real-world feedback (see Suppl. Sec.~\ref{subsec:replanning}), and future work will explore tighter integration of real-time feedback for improved robustness.

%% file: sec/5_conclusion.tex
\section{Conclusion}

We introduce \textbf{SIMPACT}, a novel action-planning framework that leverages simulation-enabled VLM to enable zero-shot robotic manipulation without any task-specific training. 
Our approach is made possible by a foundation-model-enabled simulation construction pipeline and a test-time VLM reasoning framework that together unlock the rich commonsense knowledge and reasoning capabilities of VLMs for physics-aware, fine-grained robotic manipulation.
Real-world experiments demonstrate that SIMPACT provides substantial improvements over state-of-the-art general-purpose manipulation models. 
Additional ablation studies further highlight the importance of both simulation construction and test-time reasoning in achieving generalizability and high performance.

\section*{Acknowledgements}
We would like to thank Sicong Pan and Hanxiao Jiang for insightful discussions on experimental design, Xiaoshen Han for assistance with hardware setup preparation, Wei-Cheng Huang and Kaifeng Zhang for valuable suggestions on real2sim implementations, and Fangchen Liu for MOKA implementation. This work was partly supported by the Kempner Institute, Amazon, and Pickle Robot.

%% file: sec/X_suppl.tex
\clearpage
\setcounter{page}{1}

\appendix
\maketitlesupplementary

This supplementary material provides additional implementation details, experiment analyses, and qualitative results supporting our main paper. We describe the full simulation-construction pipeline, including VLM-based prediction of rigid and deformable object parameters, as well as the symbolic action space and prompting strategy used for action optimization.

Additionally, we present more qualitative examples, an ablation on the number of VLM-sampled action proposals, and a study comparing a CEM-based Prompting-with-the-Future-style variant~\cite{ning2025prompting}, which shows near-zero success and highlights the importance of both the VLM sampler and VLM optimizer. We also show that \textbf{SIMPACT} demonstrates robustness under randomized scene variations, and provide representative failure cases. 

Importantly, we perform an additional experiment that analyzes the consistency between simulation and real-world performance, showing strong alignment (89\% agreement) while noting remaining sim-real gaps. 
This indicates that our VLM-Simulation integration serves as a high-fidelity world model for planning.

\section{Implementation Details}

\subsection{Simulation Construction Details}
\label{subsec:sim_construct}

\paragraph{Physical Parameters $\theta_{\text{phys}}$ Prediction}
In Sec.~\ref{sec:method}, we outlined the VLM-based physical parameter prediction of $\theta_{\text{phys}}$. 
Here we provide further details on how to predict the physical parameters of both rigid and deformable objects. 
This process follows a question-answering framework, where each question has the scene RGB image as an additional input to the VLM.

\begin{enumerate}
    \item[\textbf{Q1.}] Identify the objects that need to be manipulated from \{\textit{task instruction}\}.  
    Determine whether to use a rigid or deformable object simulator based on the object's material.

    \item[\textbf{Q2.}] For rigid \{\textit{object}\}, predict its mass and friction parameters.

    \item[\textbf{Q3.}] For deformable \{\textit{object}\}, decide whether to use Projective Dynamics or the Material Point Method based on the stiffness of the object.

    \item[\textbf{Q4.}] For \{\textit{object}\} simulated with Projective Dynamics, determine the following physical parameters:
    \begin{itemize}
        \item Young's modulus
        \item Poisson's ratio
        \item Mass density
    \end{itemize}

    \item[\textbf{Q5.}] For \{\textit{object}\} simulated with the Material Point Method, first determine the material type of \{\textit{object}\} from the set: \{\textit{jelly},\ \textit{metal},\ \textit{sand},\ \textit{foam},\ \textit{plasticine}\}. Then output:
    \begin{itemize}
        \item Young's modulus
        \item Poisson's ratio
        \item Mass density
        \item (Optional) Friction angle
        \item (Optional) Yield stress
    \end{itemize}
\end{enumerate}

\subsection{Action Planning Details}
\label{subsec:action_details}
This section provides further details on how our action planning framework is instantiated.

\topic{Symbolic Actions}
Here we provide a complete list of high-level symbolic actions $\textsc{a}_t$ and their corresponding continuous parameters, which are used by $\mathrm{VLM}$ in the $\textsc{Sample}$ function in Alg.~\ref{alg:planning} and Eq.~\ref{eq:sample_action}.

\begin{itemize}
    \item \texttt{PUSH}($\delta_x$, $\delta_y$): Move the end-effector horizontally from its current position by $(\delta_x, \delta_y)$ while maintaining its current height.
    \item \texttt{LIFT}($\delta_z$): Move the end-effector upwards along the $z$-axis by $\delta_z$.
     \item \texttt{DESCEND}($\delta_z$): Move the end-effector downwards along the $z$-axis by $\delta_z$.
    \item \texttt{GRASP}($d$): Adjust gripper to a target width $d$ (in meters), where $0.0$ is fully closed and $0.1$ is fully open.
    \item \texttt{RELEASE}: Fully open the gripper.
    \item \texttt{ROTATE}($\delta_{\text{yaw}}$): Adjust end-effector yaw relative to its current orientation by $\delta_{\text{yaw}}$ (in radians).
    \item \texttt{MOVE}($\delta_x$, $\delta_y$, $\delta_z$): Move the end-effector from current position by ($\delta_x$, $\delta_y$, $\delta_z$).
\end{itemize}
\label{list:primitives}

Note that these symbolic actions are redundant; for example, both the \texttt{DESCEND} and \texttt{LIFT} actions move along the $z$-axis, differing only in direction. 
However, we empirically found that this additional semantic structure allows the VLM to reason more effectively. 
For instance, in the \textit{bowl stacking} task, the VLM more reliably infers that it should descend first and then lift the bowl after grasping.

\topic{Action Proposals Generation}
Here we provide further details of $\ell_{\text{sample}}$, specifically regarding how we leverage the VLM sampler to generate action proposals. Figure~\ref{fig:l_sample} illustrates the prompt used for generating $\textsc{a}^i = \textrm{VLM}(I_0, \ell_{\text{task}}, s_0; \ell_{\text{sample}})$ as described in Eq.~\ref{eq:sample_action}.

\begin{figure*}[t]
\vspace{10px}
\centering
\begin{tcolorbox}[colback=white, colframe=black!20, width=\textwidth, left=4pt, right=4pt, top=4pt, bottom=4pt]
\textbf{Task Specification}
You are a versatile, general-purpose AI assistant functioning as an embodied planner for a robot arm. Your objective is to decompose a high-level natural language instruction into multiple distinct, high-level action plans. 
Analyze the user's instruction and scene context to propose $\#$ different, plausible action plans, each composed of a sequence of action primitives exploring different strategies.
Determine if the task requires a single primitive or a sequence of primitives. Avoid aggressive or risky proposals and focus on plans with high success rates.

\vspace{4pt}
\textbf{Input Specification}
\begin{itemize}
    \item Image of the Scene: Visual observation of the workspace.
    \item Additional Scene Context: Object and end-effector coordinates in the world frame, workspace constraints.
    \item Natural Language Instruction: High-level task goal.
\end{itemize}

\vspace{6pt}
\textbf{Action Primitive Definitions}
All coordinates $(x, y, z)$ are in the absolute world frame. The available primitives are described textually the same as list~\ref{list:primitives}.

\vspace{6pt}
\textbf{Output Specification}
Return a JSON object with key \texttt{"action\_proposals"} containing \# entries, each with:
\begin{itemize}
    \item \texttt{"description"}: Brief explanation of the high-level plan logic.
    \item \texttt{"action\_sequence"}: List of action primitives in one of the formats below.
\end{itemize}
\vspace{-5px}
\begin{verbatim}
PUSH:    {"type":"PUSH", "delta_x":float, "delta_y":float, "reasoning":"..."}
LIFT:    {"type":"LIFT", "delta_z":float, "reasoning":"..."}
DESCEND: {"type":"DESCEND", "delta_z":float, "reasoning":"..."}
GRASP:   {"type":"GRASP", "width":float, "reasoning":"..."}
RELEASE: {"type":"RELEASE", "reasoning":"..."}
ROTATE:  {"type":"ROTATE", "delta_yaw":float, "reasoning":"..."}
MOVE:    {"type":"MOVE", "delta_x":float, "delta_y":float, "delta_z":float, 
          "reasoning":"..."}
\end{verbatim}
\vspace{-5px}
\end{tcolorbox}
\caption{
    \textbf{Action sampling prompt $\ell_{\text{sample}}$ outline.} This prompt includes task specifications, input requirements, action primitive definitions, planning guidelines, and output format. It is combined with visual observations and scene context as input to the VLM sampler to generate diverse action sequence proposals. Symbol $\#$ indicates the number of proposals to generate for each call.}
\label{fig:l_sample}
\end{figure*}

\topic{Optimization Context $c$ Generation}
To instantiate the $\textsc{Optimize}$ function, we construct the context $c^i$ from the action sequence $a^i$ and the simulated state rollout $s^i$. 
We sample the state at the end of each symbolic action, where each action specifies the gripper’s Cartesian position $(x,y,z)$ and orientation (roll, pitch, yaw). For rigid objects, the numerical state consists of their full 6-DoF rigid transformation. 
For deformable objects, the numerical state includes voxel-downsampled keypoint coordinates together with the 3D bounding box of the object's point set.
Here we further provide an example context in Fig.~\ref{fig:rollout_context_example}.

\begin{figure}[t]
\vspace{6px}
\centering
\begin{tcolorbox}[
  colback=white,
  colframe=black!20,
  width=0.95\columnwidth,
  left=3pt,
  right=3pt,
  top=3pt,
  bottom=3pt,
  boxsep=2pt
]

\textbf{Example Rollout Context}

\vspace{2pt}
{\small
\begin{Verbatim}[breaklines=true, breakanywhere=true]
{
  "timestamp": "20260112_224423",
  "object_names": ["brown_box", "pocky_box"], % total 2 objects
  "waypoints": [
    {
      "position": [0.4199, -0.2452, 0.3555],
      "orientation": [0.00, 0.71, 0.70, 0.00],
      "gripper_width": 0.1,
      "duration": 3.0
    },
    ...
  ], % total 5 waypoints
  "snapshots": [
    {
      "waypoint_index": 0,
      "gripper": {
        "position": [0.4201, -0.2460, 0.2503],
        "orientation": [0.00, -0.71, -0.70, 0.00],
        "width": 0.04
      },
      "objects": {
        "pink_blue_box": {
          "position": [0.4654, 0.2061, 0.1240],
          "orientation": [-0.49, -0.47, 0.52, 0.52]
        },
        "pocky_box": {
          "position": [0.4408, -0.1004, 0.1471],
          "orientation": [0.70, 0.71, -0.02, -0.02]
        }
      },
      "screenshot": "rollout_screenshot_0.png"
    },
    ...
  ] % total 5 snapshots
}
\end{Verbatim}
}

\end{tcolorbox}
\caption{
\textbf{Example rollout context for action optimization in pivoting task.}
The context contains the action waypoints and the simulated state snapshots at each waypoint, including gripper pose, object poses, and screenshot paths.
Only the first entry is shown for repeated fields, with omitted entries summarized using comments.}
\label{fig:rollout_context_example}
\end{figure}

\begin{figure}[h!]
    \centering
    \includegraphics[trim={2.5cm 11.8cm 8.5cm 12cm},clip,width=\linewidth]
    {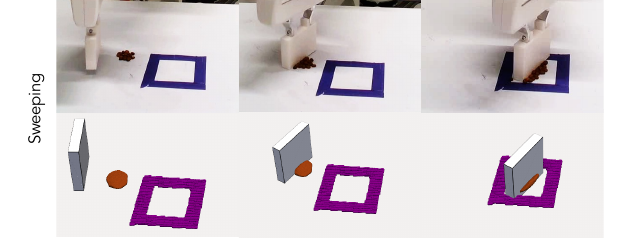}
    \caption{\textbf{Additional qualitative results.} Following Fig.~\ref{fig:qualitative}, this figure shows the initial state, execution progress, and final state for the \emph{sweeping} tasks.}
    \label{fig:qualitative_remaining}
\end{figure}

\topic{Action Optimization}
We provide details the action optimization prompt $\ell_{\text{opt}}$ in Fig.~\ref{fig:l_opt}, which enables a VLM to serve as an action optimizer. 
The prompt includes three key elements: task specification, input specification, and output specification. 

\begin{figure*}[t]
\vspace{10px}
\centering
\begin{tcolorbox}[colback=white, colframe=black!20, width=\textwidth, left=4pt, right=4pt, top=4pt, bottom=4pt]

\textbf{Task Specification}

You are an AI assistant acting as an embodied planner. Your objective is to analyze simulation rollouts and propose one optimized action plan for a real-world task. Simulation and real-world physics are similar but may differ due to sim2real gaps (e.g. appearance, pose, scale, friction).

1) Analyze Rollouts: Inspect each rollout’s \texttt{action\_sequence}, robot/object poses at each waypoint, and screenshots.

2) Infer Logic \& Physics: Identify the causes of failures and the characteristics of successful attempts.

3) Propose Optimized Plan: Output a refined plan that avoids prior mistakes and leverages successful rollout elements.

\vspace{4pt}

\textbf{Input Specification}

\begin{itemize}
    \item Task Instruction: Main task goal.
    \item Real-World Context: Workspace limits, safe ranges
    \item Simulation Rollouts: Specify the format of input context describing action and state. 
\end{itemize}

Screenshots appear as \texttt{fig\_0.png}, \texttt{fig\_1.png}, ... and should be used to evaluate rollout outcomes.

\vspace{6pt}

\textbf{Output Specification}

Produce a JSON object with key \texttt{"action\_proposals"} containing exactly one entry:
\begin{itemize}
    \item \texttt{"description"}: How the new plan improves on the rollouts.
    \item \texttt{"action\_sequence"}: A list of actions in one of the formats below.
\end{itemize}

\vspace{-5px}
\begin{verbatim}
Move Action: {
  "type":"move", "delta_x":float, "delta_y":float, "delta_z":float,
  "delta_roll":float, "delta_pitch":float, "delta_yaw":float, "reasoning":"..."
}
Gripper Control: {
  "type":"gripper_control", "width":float, "reasoning":"..."
}
\end{verbatim}
\vspace{-5px}

\end{tcolorbox}
\caption{
    \textbf{Action optimization prompt $\ell_{\text{opt}}$ outline.} 
    This prompt includes task, input, and output specifications. 
    It is combined with simulation rollout context as input to the VLM action optimizer to generate optimized action sequences.}
\label{fig:l_opt}
\end{figure*}

\section{Supplementary Results}

\subsection{Additional Qualitative Results}
\label{sec:additional_qualitative_results}
We show qualitative results for the \emph{sweeping} task that was not included in the main paper due to space constraints in Fig.~\ref{fig:qualitative_remaining}.

\subsection{Further Ablation Analysis}

We additionally consider a variant of our method in which we simultaneously replace the VLM sampler with a random sampler and switch the VLM optimizer to a sampling-based optimizer (e.g., the cross-entropy method). 
In this setting, the VLM serves only as an evaluator used to select the best rollout.
Notably, this simplified variant is algorithmically identical to Prompting-with-the-Future (PWTF)~\cite{ning2025prompting}.

We follow the open-sourced CEM implementation from PWTF and adopt the same set of hyperparameters. 
We evaluate this variant and find that it consistently achieves a zero success rate across all of our real-world tasks. 
This result further highlights the importance of both the VLM sampler and the VLM optimizer, as a naive initial sampling distribution combined with a local update process has limited performance. 
The original CEM optimization appears effective in PWTF primarily because their experiments focus on pick-and-place problems, for which sampling a reasonable initial solution is relatively easy.

\subsection{Correlation Between Simulation and Real-World Performance}

\begin{figure}
    \centering
    \includegraphics[trim={3.cm 11cm 1.5cm 9cm},clip,width=\linewidth]
    {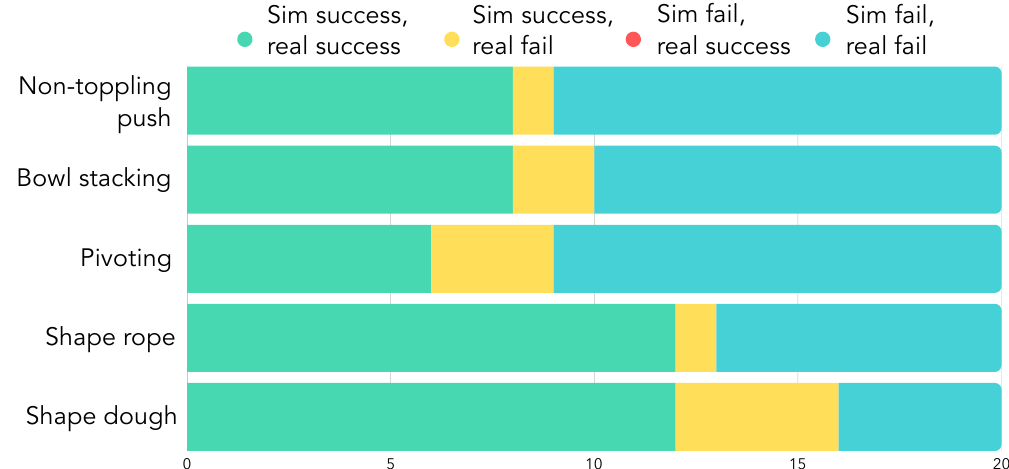}
    \vspace{-4mm}
    \caption{
    \textbf{Correlation Between Simulation and Real-world Success/Failure.} 
    Results from 20 samples per task (100 total). Each rollout is categorized as one of: sim-success/real-success (green), sim-success/real-fail (yellow), sim-fail/real-success (red), and sim-fail/real-fail (blue). 
    Simulation and real outcomes match in 89\% of cases (both success or both failure), with 11\% showing sim-success/real-fail.
    We observed no cases where a sequence failed in simulation but succeeded in the real world.
    }
    \label{fig:sim2realgap}
    \vspace{-5pt}
\end{figure}

This section examines the correlation between simulation and real-world results, specifically whether success or failure in simulation predicts the corresponding real-world outcome.
Since our planning pipeline optimizes action sequences for task success, we also include 10 unoptimized VLM sampled action proposals to capture failure cases to better understand the sim-to-real gap. 
We conduct this experiment on five selected tasks \emph{non-toppling}, \emph{bowl stacking}, \emph{pivoting}, \emph{shape rope} and \emph{shape dough}.
Each task therefore has 20 samples: 10 from the main experiments using our full pipeline, and 10 using direct VLM sampled action sequences. Results are shown in Fig.~\ref{fig:sim2realgap}. 

Across tasks, we observe a high degree of consistency between simulation and real-world outcomes, with 89\% of all cases exhibiting aligned success or failure. 
Such alignment is critical to the effectiveness of our approach, as it indicates that the physical simulation provides a reliable grounding for VLM-based planning. 
Simulated failures enable the VLM to avoid similar real-world failures, while simulated successes offer informative guidance for selecting effective action sequences. 
Despite the overall high alignment ratio, there remains room to improve simulation and real consistency. 
In the \emph{pivoting} task, 15\% of cases succeed in simulation but fail in the real world, and in the \emph{shape dough} task this discrepancy ratio is 20\%.
These tasks appear more sensitive to accurate physical modeling and contact dynamics. 
Improving simulation fidelity, e.g., via system identification, may reduce these discrepancies and prevent our planner from selecting actions that succeed in simulation but fail in the real world.

\subsection{Computation Time}
Table~\ref{tab:compute} reports the runtime of each component in our method. 
We observe that the simulation construction stage takes less than two minutes on average, with the majority of the time consumed by running the pretrained image-to-3D model. The image segmentation and pose estimation steps require significantly less time.

The VLM planning stage is the most time-consuming component. 
This is primarily due to the reasoning time of the VLM as well as the network latency introduced by querying the Gemini API. 
Within this stage, the largest portion of the runtime comes from action sampling. 
This is because we intentionally perform multiple VLM queries to encourage diversity in the generated action proposals, rather than relying on a single query to produce all samples. 
With more efficient VLMs tailored for robotics applications, the planning loop could be made substantially faster.

The total simulation stage takes less than one minute on average.
Our physical simulation has been optimized for efficiency, where each rollout lasting 5--8 seconds depending on the task. 
Implementing batched simulation for multiple rollouts would further reduce the overall simulation time.

\begin{table}[t]
\centering
\caption{\textbf{Computation time.} We compute the average computation time over 10 cases from each task.}
\label{tab:compute}
\vspace{-5pt}
\small
\begin{tabular}{lcc}
\toprule
Component & Time (mins) \\
\midrule
simulation construction & 1.9 \\
action sampling & 2.8 \\
simulation rollout & 0.8 \\
action optimization & 0.9 \\
\bottomrule
\end{tabular}
\vspace{-0.1em}
\end{table}

\subsection{Robustness Validation}

\begin{figure}[h]
    \centering
    \includegraphics[trim={3.3cm 13cm 3.5cm 10cm},clip,width=\linewidth]
    {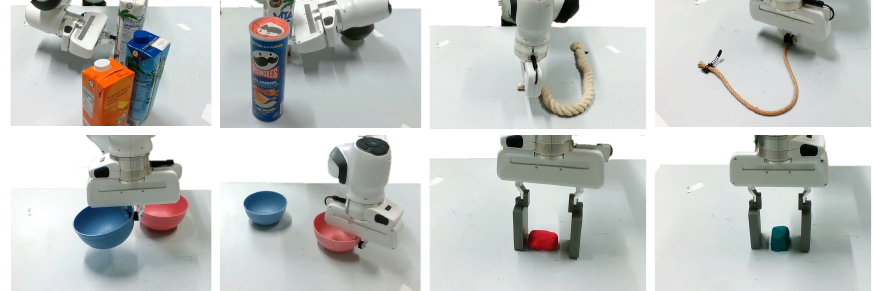}
    \vspace{-1em}
    \caption{
    \textbf{Example scene setup variations.} 
    Throughout our experiments, we vary the object types, poses, colors and materials to demonstrate the robustness and generalizability of our method.   
    }
    \label{fig:variations}
    \vspace{-1pt}
\end{figure}

We validate the robustness of our method by randomizing the scene layout and introducing different distractors for each rollout, as illustrated in Fig.~\ref{fig:variations}. Our evaluation highlights robustness across several dimensions, including the presence of unrelated objects in the environment, variations in the relative positions of task-relevant objects, and changes in the color or texture of the manipulated items. These results demonstrate that our method naturally generalizes to a wide range of scene variations, owing to the strong scene-understanding capabilities of the VLM.

\begin{figure}[t]
    \centering
    \includegraphics[trim={3.cm 13cm 2.8cm 12cm},clip,width=\linewidth]
    {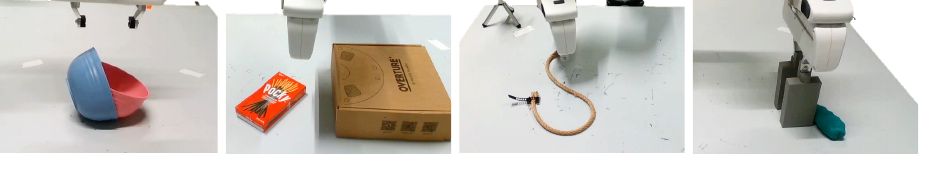}
    \caption{
    \textbf{Failure cases.} Example failure cases in bowl stacking, pivoting, shape rope and shape dough tasks. 
    }
    \vspace{-0.2em}
    \label{fig:failures}
\end{figure}

\subsection{Physics Parameters Estimation Analysis}
Accurate parameter estimation is crucial for dynamics-dependent tasks (e.g., \textit{non-toppling push}), though less significant for standard pick-and-place (e.g., \textit{bowl stacking}).
Empirically, VLM-sampled physical parameters exhibit low variance within valid ranges, as shown in Table~\ref{tab:phys_params}, likely due to VLM's rich pre-trained knowledge.

\begin{table}[H]
\centering
\caption{Robustness analysis of VLM-estimated physics parameters ($N=10$ samples). The low variance and stable ranges indicate consistent estimation capabilities.}
\label{tab:phys_params}
\resizebox{\linewidth}{!}{%
\begin{tabular}{llcc}
\toprule
\textbf{Task} & \textbf{Parameter} & \textbf{Mean $\pm$ Std} & \textbf{Range [Min, Max]} \\
\midrule

\multirow{2}{*}{\textit{Non-toppling Push}} & Mass (kg) & $1.033 \pm 0.0015$ & $[1.0, 1.05]$ \\
 & Friction Coeff. $\mu$ & $0.36 \pm 0.11$ & $[0.3, 0.5]$ \\

\midrule

\multirow{2}{*}{\textit{Shape Playdoh}} 
& Poisson's Ratio & $0.43 \pm 0.02$ & $[0.40, 0.45]$  \\
& Mass Density (kg/m$^3$) & $1186 \pm 65$ & $[1000, 1260]$ \\
 
\bottomrule
\end{tabular}%
}
\end{table}

\begin{figure*}[t!]
    \centering
    \includegraphics[trim={1.5cm 18cm 1.5cm 5cm},clip,width=\linewidth]{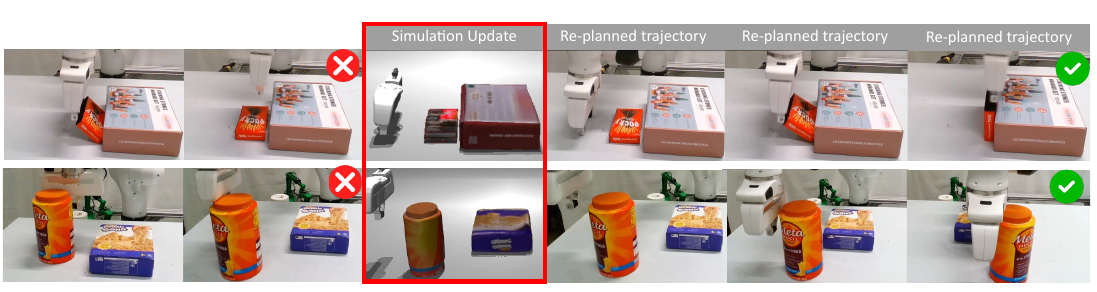}
    \caption{\textbf{Replanning illustration.} After initial failed execution, we perform re-planning after simulation update leading to successful completion. }
    \label{fig:replan}
\end{figure*}

To further verify robustness under inaccurately estimated physical parameters, we varied the object's friction coefficient $\mu$ in the \textit{non-toppling push} task. The performance degrades at physical extremes, as shown in Table~\ref{tab:friction_ablation}, where overly large friction prevents all movements, while small friction prevents toppling at all heights. However, these values all fall outside the VLM's predicted range of $[0.3,0.5]$, confirming our method's robustness to estimation errors within a valid range.

\begin{table}[h]
\centering
\vspace{-0.5em}
\caption{Success rate vs friction coefficient $\mu$.}
\label{tab:friction_ablation}
\resizebox{\linewidth}{!}{
\begin{tabular}{lccccc}
\toprule
Fric. Coeff. ($\mu$) & 0.01 & 0.20 & 0.50 & 1.00 & 10.0 \\
\midrule
Success Rate (\%) & 40   & 100  & 80   & 80 & 0 \\
\bottomrule
\end{tabular}
}
\vspace{-0.8em}
\end{table}

\subsection{Standardized Benchmark Results}

We also provide evaluation of our method on the CALVIN benchmark~\cite{mees2022calvin} containing long-horizon tasks in simulation, as shown in Table~\ref{tab:calvin_results}. We evaluated 40 chains of instructions, and used ground-truth segmentation masks given simulated environment.  Our zero-shot method outperforms imitation learning baseline HULC~\cite{mees2022hulc} and VLA baseline RoboFlamingo~\cite{li2023vision}. We also include results from the current best performing baseline FLOWER~\cite{reuss2025flower} as a reference.

\begin{table}[H]
\centering
\vspace{-0.3em}
\scriptsize
\caption{Evaluation results on the CALVIN Long-Horizon Multi-Task Language Control (LH-MTLC) benchmark.}
\label{tab:calvin_results}
\setlength{\tabcolsep}{2pt}
\begin{tabular}{lccccccc}
\toprule
\#. Instruct. & zero-shot & \textbf{1} & \textbf{2} & \textbf{3} & \textbf{4} & \textbf{5} &  Avg. Len. \\
\midrule
HULC~\cite{mees2022hulc} & \xmark & 41.8\% & 16.5\% & 5.7\% & 1.9\% & 1.1\% & 0.67 \\
RoboFlamingo~\cite{li2023vision} & \xmark & 82.4\% & 61.9\% & 46.6\% & 33.1\% & 23.5\% & 2.47 \\
FLOWER~\cite{reuss2025flower}  & \xmark & 99.4\% & 95.8\% & 90.7\% & 84.9\% & 77.8\% & 4.53 \\
\midrule
Ours & \cmark & 87.5\% & 82.5\% & 47.5\% & 40.0\% & 20.0\% & 2.78 \\
\bottomrule
\end{tabular}
\vspace{-0.2em}
\end{table}

\subsection{Failure Cases}
We present representative failure cases of our method in Fig.~\ref{fig:failures}, providing supplementary material for Sec.~\ref{sec:failure_analysis}.

The \emph{bowl stacking} and \emph{shape dough} failures are both execution failures.  
A slight misalignment during bowl placement can cause the bowl to flip, and a small offset between the gripper center and the dough center can lead to unsuccessful squeezing.  
These execution failures highlight the sensitivity and difficulty of our tasks: even minor errors in the planned actions can lead to failure.

The \emph{pivoting} and \emph{shape rope} failures are both planning failures.  
For the pivoting task, stabilizing the object upright requires solutions within a narrow range of feasible angles; when the planned angle is suboptimal, the object cannot maintain stable contact with the environment.  
Planning failures in simulation also transfer to the real world, further reducing success rates.
For the rope shaping case, we observe that failure often arises from insufficient diversity in the generated action proposals, which limits the VLM action optimizer’s ability to identify effective actions.  
Increasing the number of sampled proposals may improve performance in such cases.

\subsection{Replanning using Real-world Feedback.}
\label{subsec:replanning}

Our framework can also incorporate real-world feedback to improve the success rate after execution failures.
When the VLM determines that the planned action sequence has not achieved the task goal, we optionally invoke a replanning mechanism that allows the system to generate a new action sequence using updated information.
More concretely, the simulation is updated with current real-world states for a new planning attempt as shown in Fig.~\ref{fig:replan}.
We evaluate this replanning mechanism on both the \textit{pivoting} and \textit{avoid obstacle} tasks, allowing up to 3 replanning attempts. For the \textit{pivoting} task, among the six initially failed cases, replanning successfully recovers 50\% of them, with an average of 1.67 replanning attempts. For the \textit{avoid obstacle} task, among the two initially failed cases, replanning resolves all of them, with an average of 1.0 replanning attempt. These results highlight the promise of incorporating real-world feedback into our system for replanning and suggest its potential in enabling robust manipulation.